\documentclass{llncs}

\usepackage[utf8]{inputenc}
\usepackage{graphicx}

\title{Automatic glissade determination through a mathematical model in electrooculographic records}
\author{Camilo Velázquez-Rodríguez\inst{1}, Rodolfo García-Bermúdez\inst{2}, Fernando Rojas-Ruiz\inst{3}, Roberto Becerra-García\inst{4} and Luis Velázquez\inst{5}}
\institute{Universidad de Holguín, Grupo de Procesamiento de Datos Biomédicos (GPDB), 80100, Holguín, Cuba \\
	\email{cvelazquezr@uho.edu.cu}
	\and Universidad Técnica de Manabí, Facultad de Ciencias Informáticas, Manta, Ecuador\\
	\email{rodgarberm@gmail.com}
	\and Universidad de Granada, Departamento de Arquitectura y Tecnología de Computadores , ETS Ing. Informática y Telecomunicación, Granada, España\\
	\email{frojas@ugr.es}
	\and Universidad de Málaga, Málaga, España\\
	\email{idertator@gmail.com}
	\and Centro para la Investigación y Rehabilitación de Ataxias Hereditarias, Holguín, Cuba\\
	\email{cirahsca2@cristal.hlg.sld.cu}
}

\begin{document}

\maketitle
\begin{abstract}
The glissadic overshoot is characterized by an unwanted type of movement known as glissades. The  glissades are a short ocular movement that describe the failure of the  neural programming of saccades to move the eyes in order to reach a specific target. In this paper we develop a procedure to determine if a specific saccade have a glissade appended to the end of it. The use of the  third partial sum of the Gauss series as mathematical model, a comparison between some specific parameters and the RMSE error are the steps made to reach this goal. Finally a machine learning algorithm is trained, returning expected responses of the presence or not of this kind of ocular movement.
\end{abstract}

\section{Introduction}
Several events are present in electrooculographic (EOG) records, classified as ocular movements  according to different criteria. The saccades and fixations are two of the most studied events due to the clinical meaning  of those signals. The saccades are fast and accurate ballistic eye movements used in repositioning the fovea to a new location in the visual environment \cite{goffart_saccadic_2009} and the fixations are present when an object of interest is held approximately stable on the ocular retina \cite{lukander_measuring_2004}.

There are various tests that can be realized in order to evaluate a certain condition of a subject,  such as saccadic test, smooth pursuit test, horizontal and vertical calibration, etc. The  saccadic  tests are those related with the capture of saccades and fixations as main events. This kind of test is developed in the presence of a stimulus that the subject had to follow with the eyes.

The EOG records have been studied through several years and by many authors. Some characteristics of this kind of biological record that can be observed without much effort are the level of noise that comes with the signal, amplitude and length of the record. Other characteristics are inherent to  this  signal, so it is necessary to apply  different techniques  of signal processing to be able of studying it, for instance, the exact points of several events, the velocity in any point of the record and the independent components responsible of the generation of events like the saccades mentioned before.

The exact determination of the events present in EOG records is a complex task due to factors such as the error in the measure of the signal, the presence of unwanted noise captured from several sources and the peculiar dynamics of a very fast kind of movement such as the oculars. However, investigations like \cite{inchingolo_identification_1985,juhola_detection_1985,wyatt_detecting_1998,salvucci_identifying_2000} propose procedures for the identification of saccades and fixations in EOG records.

The work developed by Becerra et. al in \cite{becerra-garcia_non_2015} reinforce the use of machine learning models that works in a very efficient way in tasks like the determination of events. The investigations mentioned previously allow the extraction of events like saccades and fixations from EOG records, even with the presence of noise. However, in the majority of saccades a phenomenon known as glissadic overshoot happens, which is disregarded and not considered in the segmentation of the signal.

The  glissadic overshoot is characterized by an unwanted type of movement known as glissades. The glissades are a short  ocular movement that describe the failure of the neural programming of saccades to move the eyes in order to reach a specific target. After the target is passed, comes a rectification  of the eyes with the  objective  to  finally reach  the  desired  goal. The  steps  mentioned above  as failure and rectification are summarized by the glissade \cite{bahill_glissades--eye_1975}.

One of the first authors that investigated this type of phenomena was Terry Bahill. In \cite{bahill_glissades--eye_1975} he hypothesized that this kind of ocular movement is generated by a mismatch between the neural components that generate the saccades, the pulse and step. Also, in \cite{bahill_overlapping_1975} he concludes that from the biological point of view, this movement appended in the end of saccades is caused by fatigue in the saccadic eye movement system. The use of computational models allow the study of overshoot in saccadic eye movements as is exposed in \cite{bahill_computer_1975}. One of the  reasons that provoke the mismatch between the neural signals could be pulse width errors as mentioned in \cite{bahill_glissadic_1978}.

\begin{figure}
	\centering
	\includegraphics[scale=.7]{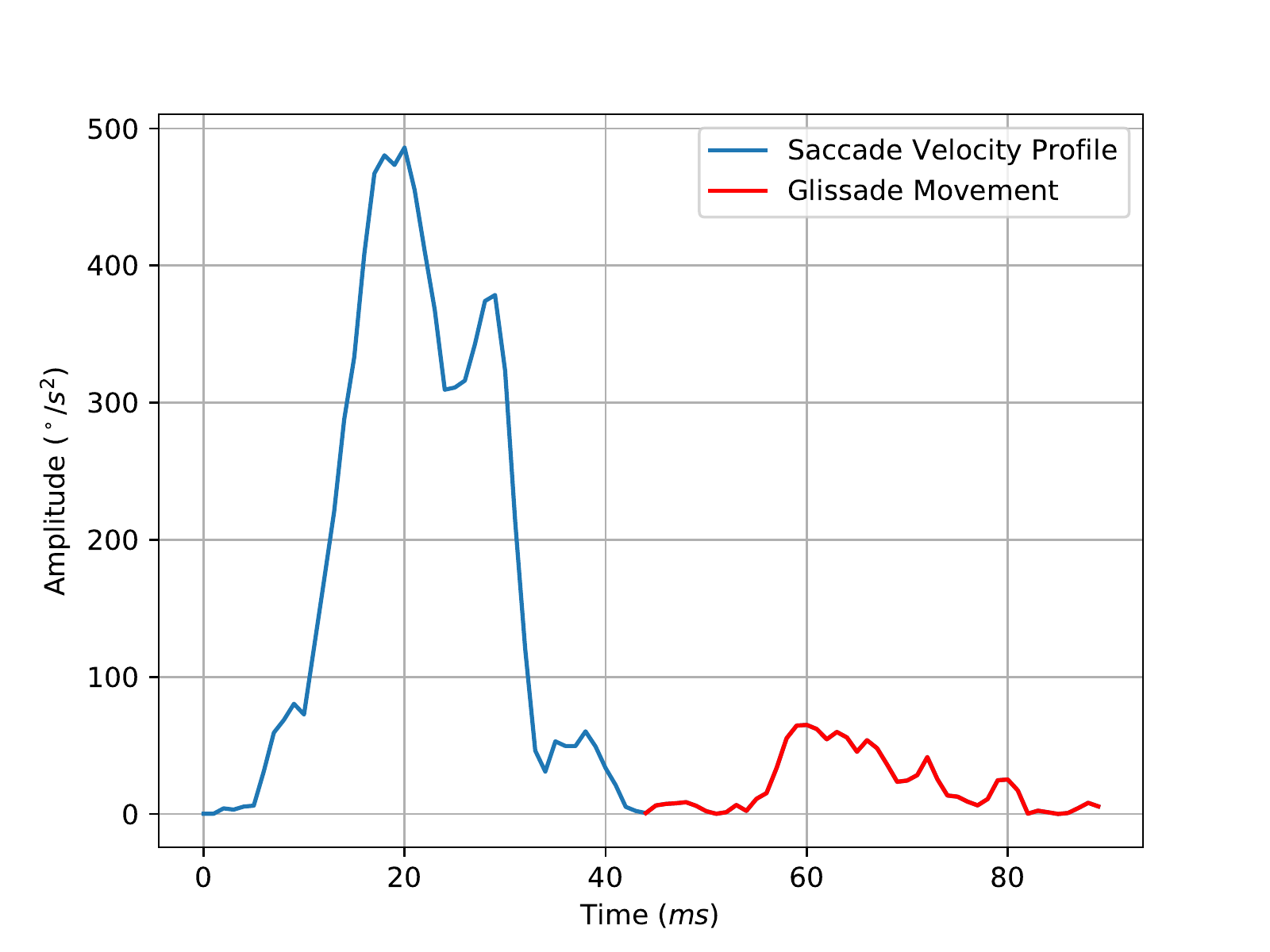}
	\caption{Numerical differentiation of a saccade with a glissade appended.}
	\label{fig:glissadic_velocity}
\end{figure}

As the glissadic overshoot is appended after the saccade finalize, sometimes it is considered  as part of the saccade movement, which provoke an extended signal larger than the standard saccadic movement. This  phenomenon  also provokes a shorter fixation caused by the extension of saccade. When a numerical differentiation to the  EOG record with glissades is applied, the velocity profile obtained shows a minor movement after the saccade, as shows the Figure \ref{fig:glissadic_velocity}.

In a previous work \cite{garcia-bermudez_evaluation_2015}, we describe a mathematical model that make a very good approximation of the velocity profile of a saccade. Through the realization of this work we use a modified version of the previous mathematical model, with the main objective of characterizing the glissadic phenomena.

The knowledge of the meaning  of the parameters in the used model, will allow us analyze those related with the glissadic phenomena. The values of these parameters and the presence or not of glissades will let the construction of a dataset in order to train a classification machine  learning algorithm. The model trained will classify new saccades with the presence or not of glissades and will give a tool for the researchers to know how to determine a better segmentation of events.

In section  2 we describe how we did the acquisition of the data, as well the mathematical model that we use to describe both, the saccade and glissade. The section 3 is dedicated to the  presentation of results and a discussion of it. Finally the conclusions resume all the work done in this investigation.

\section{Materials y Methods}

The  electrooculographic records were obtained by the  medical staff of the  Centre  for the  Research  and  Rehabilitation of the  Hereditary Ataxias  (CIRAH) at Holguín, Cuba. A two-channel electronystagmograph (Otoscreen, Jaeger-Toennies, Hochberg, Germany) was used to record saccadic ocular movements. The stimulus and patient response data  were automatically stored in ASCII files as comma separated values (CSV) by the Otoscreen  electronystagmograph, according to its user manual specifications.

Electrooculographic study from several subjects were captured at a sampling rate of 200 Hz with a bandwith of 0.02 to 70 Hz, for a total of 163 study  made. Each one of the study have at least tests of 10$^\circ$, 20$^\circ$ and 30$^\circ$ of visual stimulation, therefore containing  several records captured to the same subject. Typically, saccadic record have at least one horizontal channel and one stimulus  signal.

After the records were captured, each one of the tests included in every record was processed. In a first step, the signals recorded were filtered in order to remove the unwanted noise that carry  many of these tests. The median filter has proven to be very robust in eliminating high frequency signal noise while preserving sharp edges. A study carried out by Juhola in 1991 demonstrated that this kind of filters is appropriate for eye movements  signals \cite{juhola_median_1991}. For eliminating non desired  noise present in the signals we use a median filter with a window size of 15 samples  obtaining good results. This is accomplished using the \textbf{medfilt} function of the scientific Python library SciPy \cite{jones_e_scipy:_2001}.

To obtain the velocity profile of the filtered signals, it was applied one of the Lanczos differentiators, specifically Lanczos 11 or Lanczos of 11 points,  due that this kind of numerical  differentiators use curve fitting instead  of interpolation in their procedure \cite{becerra_garcia_plataforma_2013}. The following equation  shows the mathematical description of Lanczos 11:

\begin{equation}
	f' \approx \frac{f_{1} - f_{-1} + 2(f_{2} - f_{-2}) + 3(f_{3} - f_{-3}) + 4(f_{4} - f_{-4}) + 5(f_{5} - f_{-5})}{110h}
	\label{equ:lanczos_11}
\end{equation}

where h is the step of sampling,  which in the obtained records has a value of $h = 4.88 ms$.

\begin{figure}
	\centering
	\includegraphics[scale=.7]{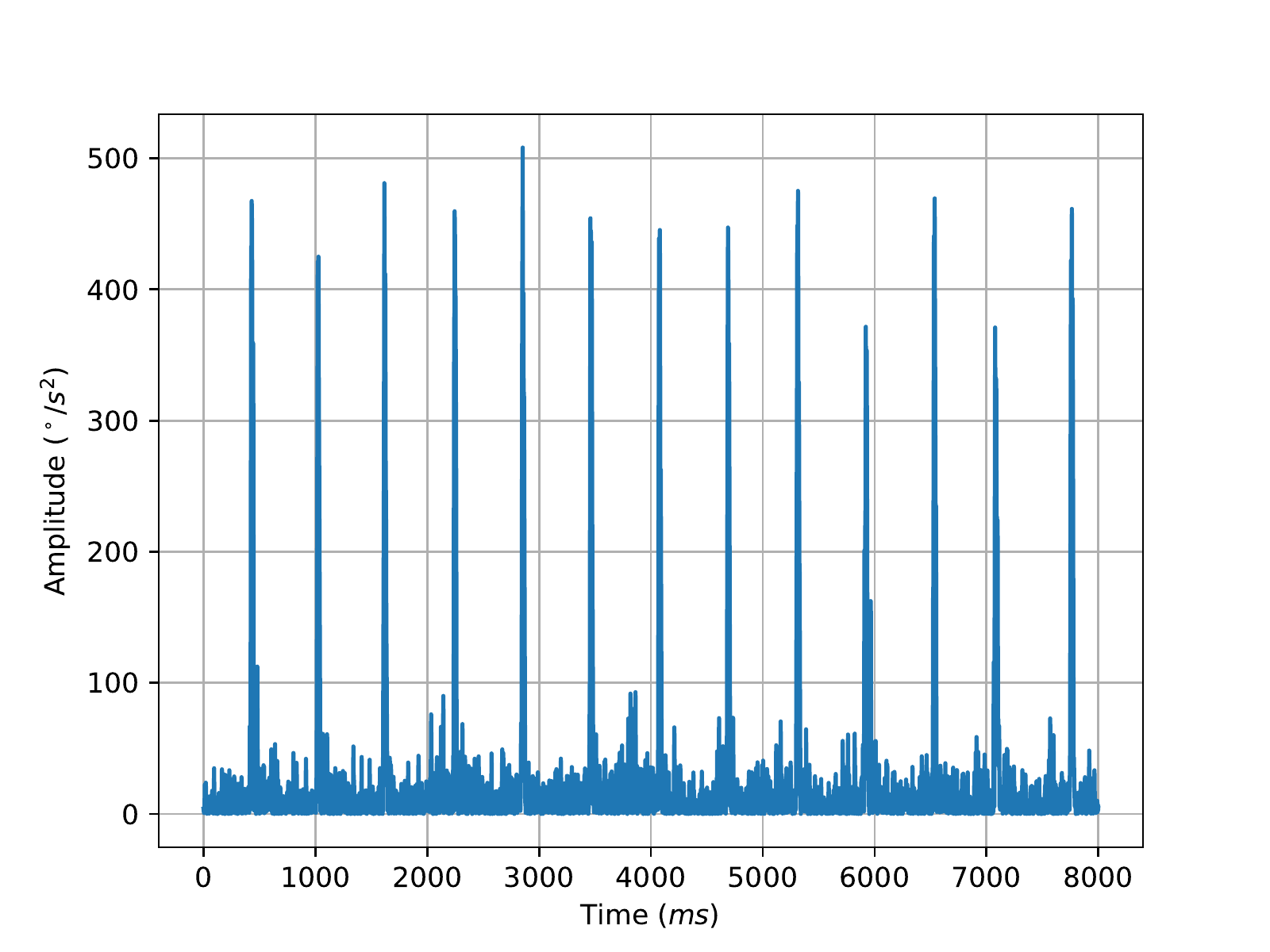}
	\caption{Velocity profile as result of Lanczos 11 and the \textbf{abs} function application.}
	\label{fig:velocity_profile}
\end{figure}

The velocity profile obtained as a result of the application of Lanczos 11, present positive and  negative values representing in a physical way the direction of the saccadic movement. In order to standardize all the velocity profiles obtained, the \textbf{absolute value} function provided  natively by Python and also included in the numerical library of this programming language named NumPy, was applied to the differentiated signal \cite{oliphant_t_python_2007}. One of the velocity profiles obtained is shown in the Figure  \ref{fig:velocity_profile}.

As can be seen in the Figure \ref{fig:velocity_profile}, there are several velocity profiles in a signal, so it is necessary to split the signals according to the different events that occur in it, in order to get every profile extracted and ready to be modeled. With the goal of separating the velocity profiles we use a \textbf{findpeaks} MATLAB similar function in the Python library. The mentioned function  is called \textbf{peakutils} and is founded in the web repository of Python located in http://pypi.python.org.

\begin{figure}
	\centering
	\includegraphics[scale=.7]{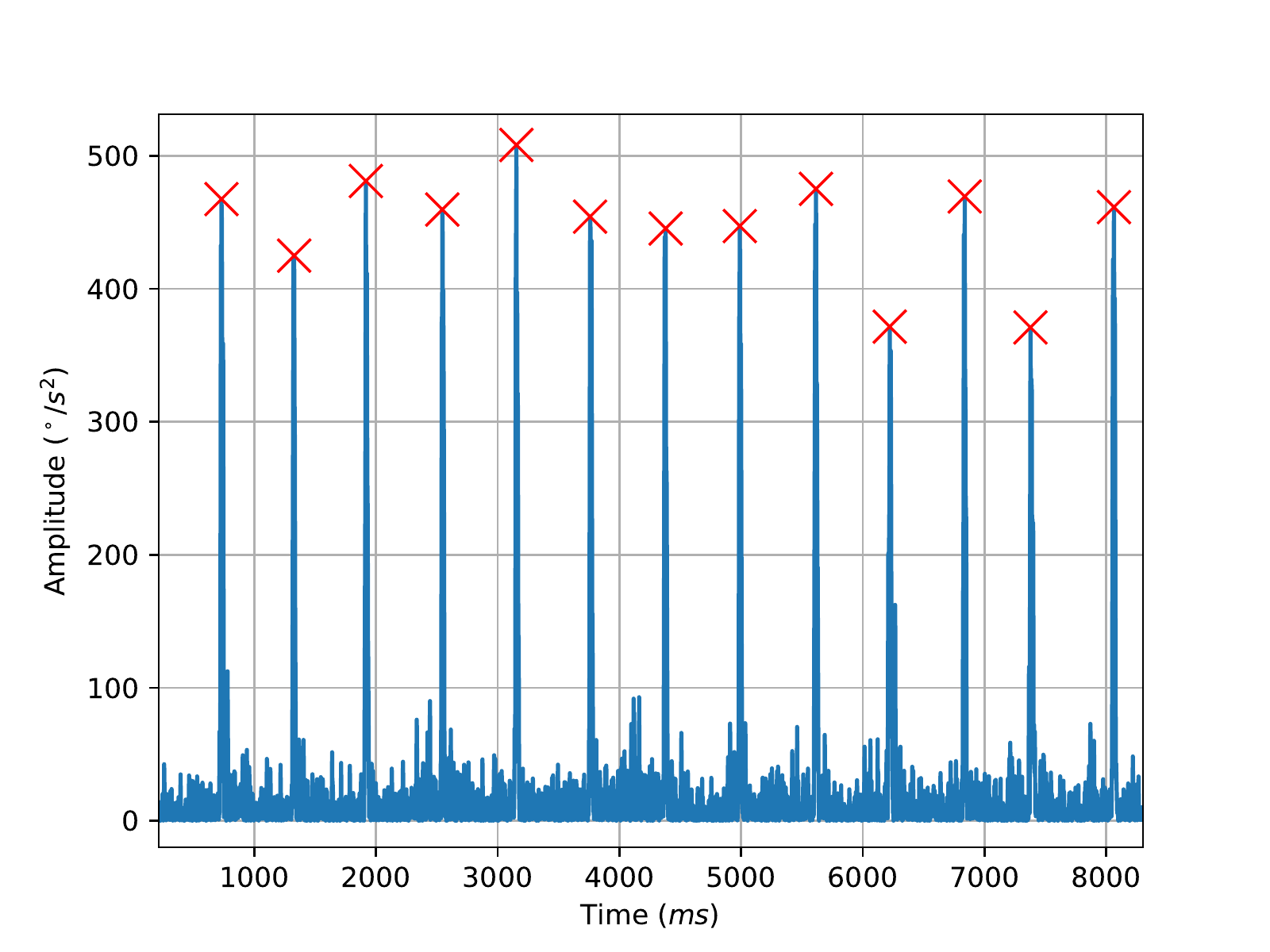}
	\caption{Peaks of the velocity profiles detected by the use of \textbf{peakutils} module.}
	\label{fig:peaks_velocity}
\end{figure}

\begin{figure}
	\centering
	\includegraphics[scale=.7]{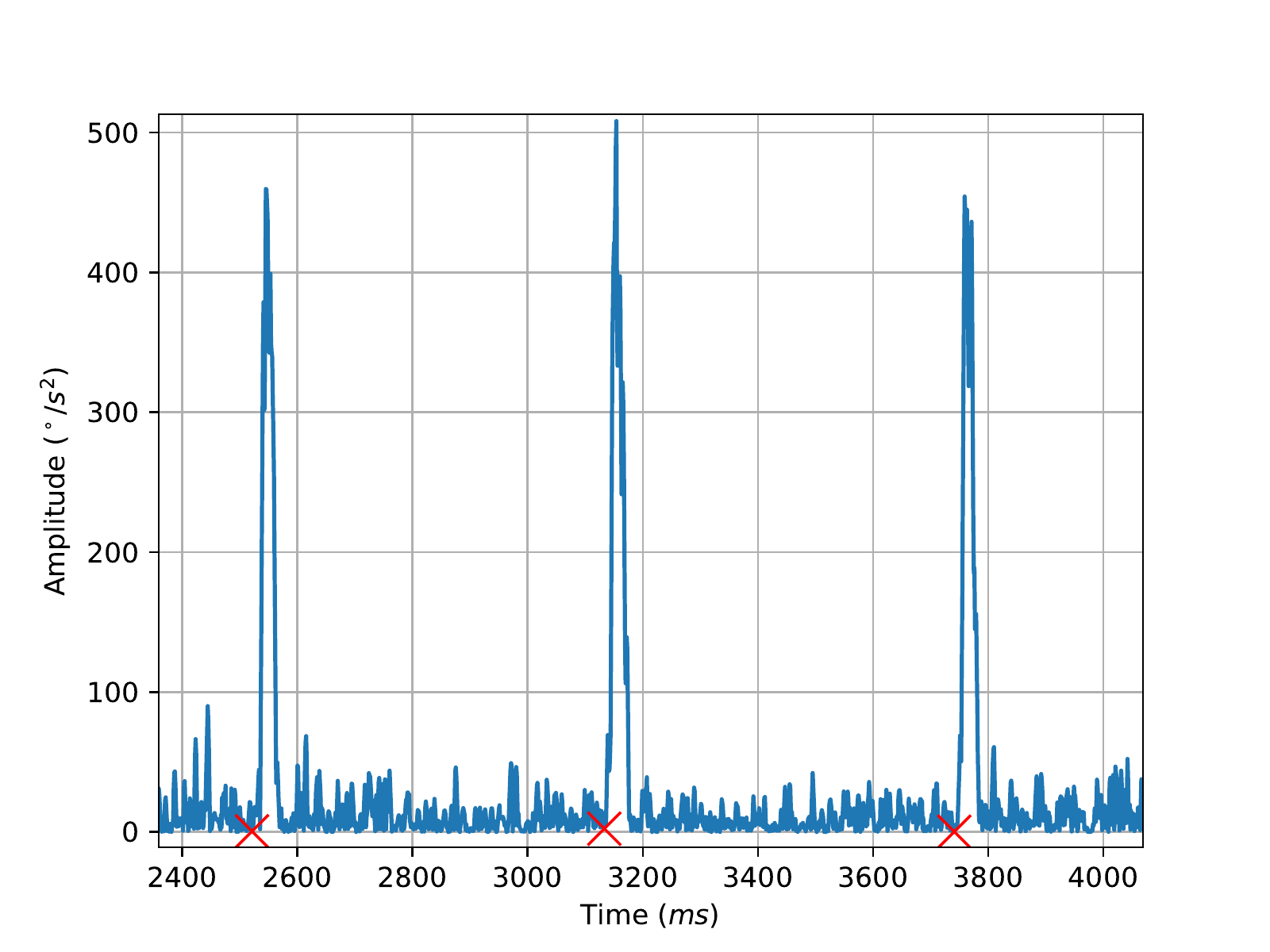}
	\caption{Start points detected using the backward procedure.}
	\label{fig:start_points}
\end{figure}

The Figure \ref{fig:peaks_velocity} shows the  peaks, marked as red exes, detected through the use of the \textbf{peakutils} library mentioned before. To find the start points we use an iterative algorithm that start in the maximum peaks detected and in every iteration go backward, after  reaching a defined threshold, we find the minimum in the  population of points near to the stopping criteria. The results of this procedure are shown in the Figure \ref{fig:start_points}.

Detected the start points for every velocity profile in the differentiated signal, we can split the latest from start point to the next start point and obtain a part of the signal similar to the one showed in the Figure \ref{fig:glissadic_velocity}. The last velocity profile is chosen until the end of the signal.

A velocity profile extracted from the differentiated electrooculographic test, could be modeled for  various equations that describe this kind of shape. In a previous work \cite{garcia-bermudez_evaluation_2015} we compare several mathematical models in the task of describing the velocity profile. The  best model in that work was the second partial sum of the Gauss series or \textbf{gauss2} for short, due to have one of the lower errors in modeling the data, only surpassed by the third partial sum of the Gauss series or \textbf{gauss3}.

We choose in \cite{garcia-bermudez_evaluation_2015} gauss2 over gauss3 because of the  presence  of 3 less parameters to be optimized and similar fitting results. However, in this  new scenario, we need to use gauss3 because not only the velocity profile will be fitted, but also the glissade signal appended in the end of the latest. The third partial sum of the Gauss series is described as follows:

\begin{equation}
	gauss3(x) = \sum_{i=1}^{3} a_ie^{\left[ -\left( \frac{x - b_i}{c_i} \right)^2 \right]}
	\label{equ:gauss3}
\end{equation}

where $a_i$ the amplitude, $b_i$ is the location of the centroid and $c_i$ is related to the width  of every gaussian peak that is optimized in order to get a good approximation of the model. The  most significant parameters for the purpose of this work are $b_1$, $b_2$ and $b_3$. These parameters give the location of centers of each gaussian peak, it is very important to analyze the behavior of this parameters because they can reflect the presence or not of glissades in a velocity profile.

The significance of the parameters involved in the mathematical modeling will present different values according to fitting made. However, if the error in the fitting is large, the meaning  of $b_i$ as the parameters taken in consideration, loose validity. The  standard error of regression,  also known as Root Mean Squared Error (RMSE) is one of the statistical metrics more used in modeling procedures. The Equation \ref{equ:rmse} gives an insight of this metric:

\begin{equation}
	RMSE = \sqrt{\frac{\displaystyle{\sum_{i=1}^{n} (y_{i} - \hat{y}_{i})^2}}{n}}
	\label{equ:rmse}
\end{equation}

where \textit{n} is the amount of points in the data, $y_{i}$ represent the $i$th point of the data  and $\hat{y}_{i}$ is the $i$th value estimated by the model in evaluation. Values closer to 0 in this metric mean that the fit is more useful for prediction.

\section{Results}

The 163 study captured, contain 904 electrooculographic records that were used to extract the velocity profiles that conform the available data  in this work. After the  application of the  mentioned procedure in the before section, we extracted 25853 velocity profiles from the EOG records. These data  were partitioned, taking 13101 velocity profiles to realize the mathematical curve fitting by the use of gauss3, representing the 50.7\% of the available  data.

\begin{figure}
	\centering
	\includegraphics[scale=.3]{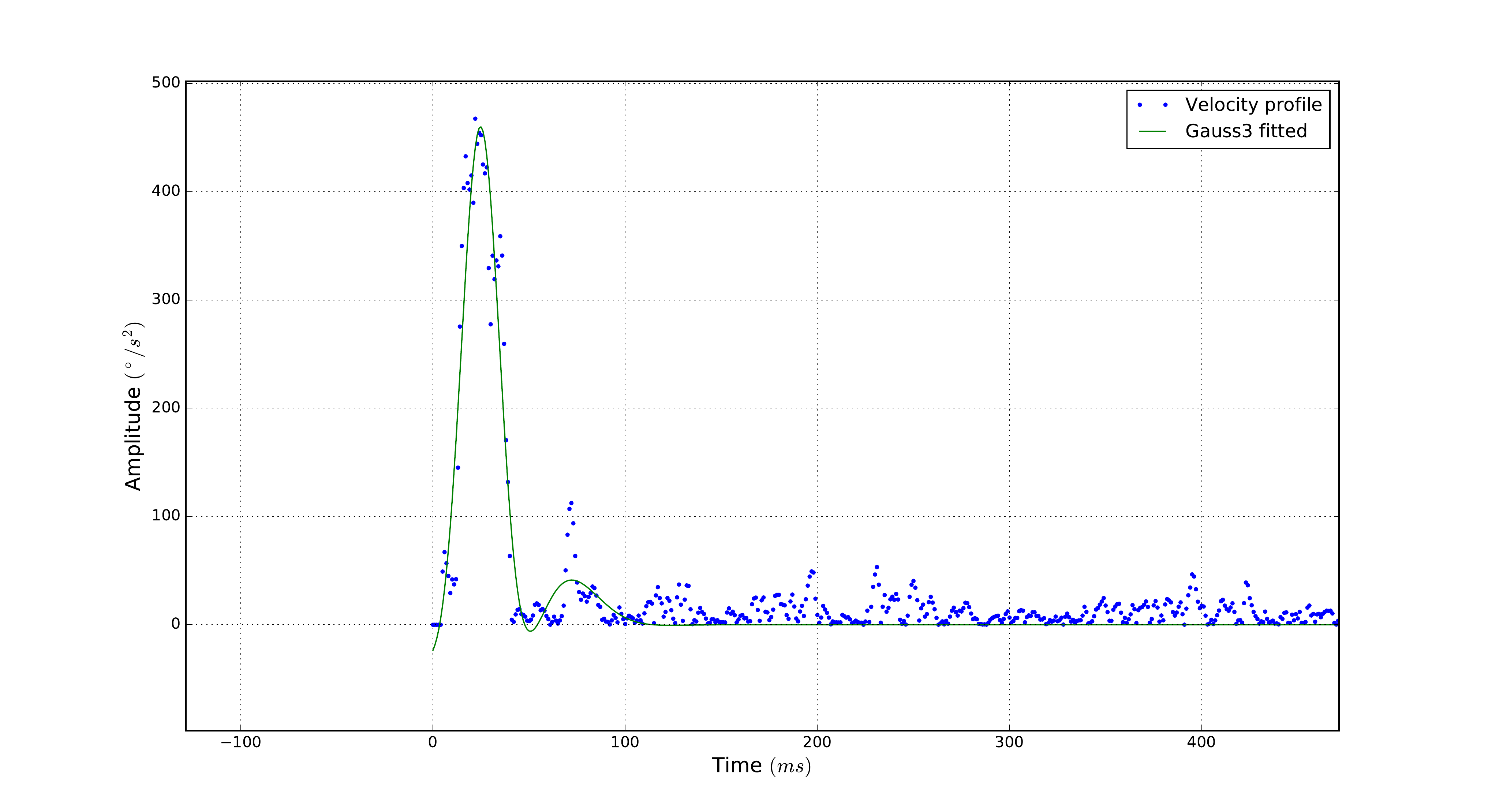}
	\caption{Velocity profile with glissade, fitted using gauss3 as model.}
	\label{fig:gauss3_fitted}
\end{figure}

As shows the Equation \ref{equ:gauss3}, the parameters $b_i$ locate the center of each one of the  gaussian peaks. We hypothesize that if the error of the curve fitting is low and the saccade present glissade at the  end of it, the gauss3 curve will fit this rectification phenomena, as shows the Figure \ref{fig:gauss3_fitted}.

The error in the curve fitting that shows the Figure \ref{fig:gauss3_fitted} was 21.8, meaning that it was not an excellent modeling, but good enough to partially describe the glissadic curve as shows the mentioned figure. This visual perception comes more clearly analyzing the chosen parameters $b_i$ with values 48.9, 24.8 and 52.8. The parameters values manifest a separation between one of the values and the other two that are very close to each other.

Making this analysis, a procedure can be formulated for determining the presence or not of glissadic phenomena, due to the values of the selected parameters. This  procedure is as follow: analyze  the values of the $b_i$ and calculate the absolute values of their differences then, inspect if the three differences are lower than a defined threshold in order to return an unmismatch response, any other case contain the presence of glissades accordingly to the gauss3 mathematical model.

With the data collected by different responses we are able to form a training dataset for a classification algorithm. This dataset will contain the RMSE error result of the fitted procedure,  each one of the parameters $b_i$ and the response of the presence or not of glissades as the class of this formed dataset.

We use machine learning tools to solve the classification task to determine if a saccade have the presence of glissades or not. Because we are using Python technologies, we selected Scikit-Learn as machine learning library, specifically we evaluate four different models: Support Vector Machines \cite{cortes_c_support-vector_1995}, CART decision trees \cite{breiman_l_classification_1984}, K-Nearest Neighbors \cite{silverman_b.w_important_1989}, and an ensemble method known as Random Forest \cite{breiman_random_2001}.

\begin{figure}
	\centering
	\includegraphics[scale=.3]{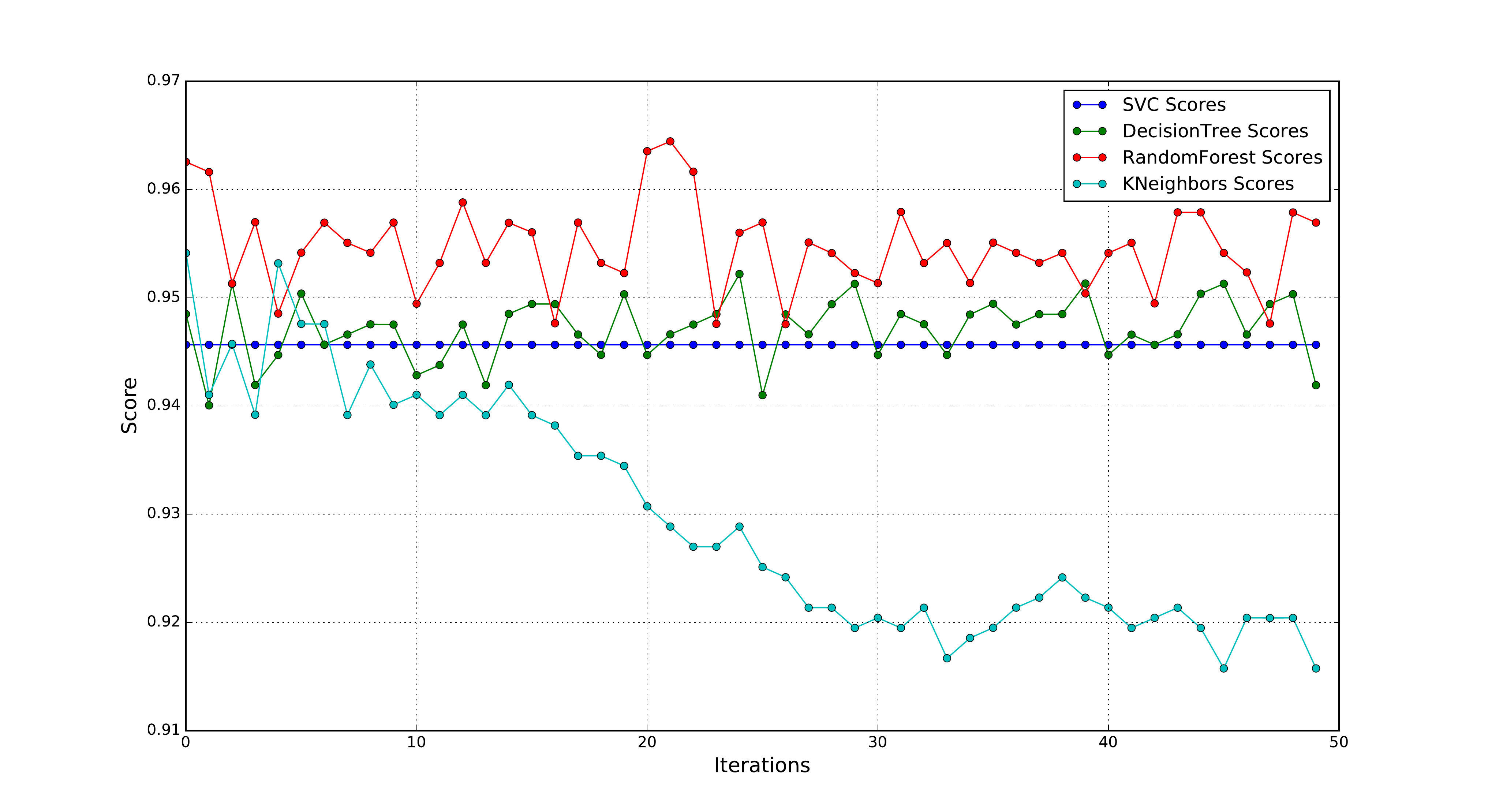}
	\caption{Classification algorithms training responses.}
	\label{fig:models_learning}
\end{figure}

Each classification algorithm was trained using a technique known as cross validation, we use the  folding equals to 10, meaning that the dataset is sliced in 10 equals parts, one for training and  the rest for evaluation, exchanging the trained part in every evaluation. As a final score of the algorithm we find the mean of trained values in each fold, also we find the standard deviation in order to know the total spectrum of the models score. With the goal of observe the behavior of each model, we apply the detailed previous procedure 50 times as can be seen in the Figure \ref{fig:models_learning}.

It is important to mention that we use the  classification algorithms with their default values, except from the K-Nearest Neighbors, being modified the number of neighbors in each iteration. Also, it can be seen in the Figure \ref{fig:models_learning} a  better general performance by the ensemble method known as Random Forest, the increment of neighbors in the K-Nearest Neighbors does not increase the performance of this method, on the other hand, make it worse. The best performance of KNN is when the number  of neighbors is 4. The CART decision tree, has a good performance also, but not so good like the Random Forest procedure. In the case of the Support Vector Machine algorithm remains invariable in the value of 94.5 of exactitude, due to the invariability of its parameters.

We choose the Random Forest classification trained model due to the presence of the highest scores in the training task, like shows the Figure \ref{fig:models_learning} some values surpass the 0.96 score, and also the majority of the scores are higher than the 0.95 value. The model trained can be used to evaluate new velocity profiles that were excluded from the training procedure.

The 49.3\% of the available velocity profiles was not considered in the training step in order to realize testing and evaluation of the machine learning model chosen by the authors. This velocity  profiles are not classified with the presence or not of glissades, so in order to evaluate the  trained model it is necessary to select a portion of the data randomly and manually, classify if it has  glissades or not. After obtaining the response of the model, it can be evaluated against the manual classification.

From the total of velocity profiles remaining, 100 records were chosen randomly, following this random process a normal distribution. A visual inspection was made in order to assign a class to the velocity profiles chosen. The objective is to know the presence or not of glissades, so a binary  class can be formed assigning the value of 1 to the first statement and 0 to the other.

A curve fitting procedure was applied to the 100 records selected in order to obtain the RMSE metric and the $b_i$ parameters mentioned before. These four collected values will serve as input to the trained Random Forest model. The response of the model was in correspondence with the expected values stated in the manual procedure. Many of the responses were correct, even in some of them, the trained model returned an answer better than the human prediction.

\section{Conclusions}

This contribution exposes the definition of a natural phenomenon known as glissadic overshoot. This kind of phenomenon is present in most of the saccadic ocular movements, specifically at the end of it, and it is supposed that fatigue is one of the main reasons for the presence of this particular movement.

Also, a computational procedure to automatically determine the glissades in a velocity profile signal is proposed. The computational algorithm involves the mathematical modeling by the use of the third partial sum of the Gauss series, due to the great similarity that this model has with the data analyzed.

The third partial sum of the Gauss series or gauss3 possesses several parameters that describe different parts  of the optimized  curve. Analyzing the $b_i$ parameters of gauss3, we can define the presence or not of the glissadic phenomena, due to the meaning of this specific parameters that are the centers of the gaussian curves. Also, the RMSE  error of each optimization made was measured, in order to know the validity of the values.

An algorithm was developed, that given certain threshold and the values of the $b_i$, can determine the presence or not of glissades. Obtained the RMSE error, the mentioned parameters and the class of the data, we built a dataset in order to train a machine learning classification algorithm. For this purpose four machine learning paradigms are compared: Support  Vector Machines, K-Nearest Neighbors, Random Forest and Classification and Regression Trees, resulting the Random Forest procedure with the better performance.

The trained model can automatically predict with great performance if a determined saccade has appended glissades or not. Many of the responses were in correspondence with the expected values and in others cases improve the human prediction made about certain decisions.

\vspace{0.5cm}
\textbf{Acknowledgements}: This work has been partially supported by the  project TIN2015-67020-P of the Spanish Ministry of Economy and Competitiveness

\bibliographystyle{ieeetr}
\bibliography{references}
\end{document}